\documentclass[sigconf]{acmart}
\AtBeginDocument{%
  \providecommand\BibTeX{{%
    \normalfont B\kern-0.5em{\scshape i\kern-0.25em b}\kern-0.8em\TeX}}}

\usepackage{amsmath,amsfonts, mathtools}

\usepackage[linesnumbered,lined,ruled,commentsnumbered]{algorithm2e}
\usepackage{etoolbox}
\let\oldnl\nl
\newcommand{\nonl}{\renewcommand{\nl}{\let\nl\oldnl}}
\makeatletter
\patchcmd{\@algocf@start}
  {-1.5em}
  {10pt}
  {}{}
\makeatother

\usepackage{graphicx}
\usepackage{multirow}
\usepackage{microtype}
\usepackage{caption}
\usepackage{footnote}
\usepackage{textcomp}
\usepackage{xcolor}
\usepackage{booktabs}
\usepackage{color}
\usepackage{hyperref}
\usepackage{threeparttable}
\usepackage{footnote}
\makesavenoteenv{tabular}
\usepackage{amsthm}
\usepackage{bm}
\usepackage{setspace}
\usepackage{caption}
\usepackage{tabularx}
\usepackage{multirow}
\allowdisplaybreaks[4]
 \usepackage[misc]{ifsym}
\usepackage{float}
\usepackage{changepage}
\usepackage{subfig}

\usepackage{progressbar}
\usepackage{makecell}
\usepackage{tabularx}
\usepackage{multicol}
\usepackage{multirow}
\usepackage{enumitem}
\newtheorem{assumption}{Assumption}

\newtheorem{theorem}{Theorem}

\newcommand\blfootnote[1]{%
  \begingroup
  \renewcommand\thefootnote{}\footnote{#1}%
  \addtocounter{footnote}{-1}%
  \endgroup
}

\copyrightyear{2022}
\acmYear{2022}
\setcopyright{acmlicensed}\acmConference[DAC '22]{Proceedings of the 59th
ACM/IEEE Design Automation Conference (DAC)}{July 10--14, 2022}{San
Francisco, CA, USA}
\acmBooktitle{Proceedings of the 59th ACM/IEEE Design Automation
Conference (DAC) (DAC '22), July 10--14, 2022, San Francisco, CA, USA}
\acmPrice{15.00}
\acmDOI{10.1145/3489517.3530417}
\acmISBN{978-1-4503-9142-9/22/07}

\begin{document}
\fancyhead{}

\title{Sign Bit is Enough: A Learning Synchronization Framework for Multi-hop All-reduce with Ultimate Compression}

\author{Feijie Wu$^{1\dagger}$, Shiqi He$^{2\dagger}$, Song Guo$^{1 *}$, Zhihao Qu$^{3*}$, Haozhao Wang$^4$, Weihua Zhuang$^5$, Jie Zhang$^1$}
\affiliation{
\institution{$^1$The Hong Kong Polytechnic University, $^2$The University of British Columbia, $^3$Hohai University, $^4$Huazhong University of Science and Technology, $^5$University of Waterloo}
\country{}
}
\renewcommand{\shortauthors}{}
\settopmatter{printacmref=false}

\begin{abstract}

Traditional one-bit compressed stochastic gradient descent can not be directly employed in multi-hop all-reduce, a widely adopted distributed training paradigm in network-intensive high-performance computing systems such as public clouds. According to our theoretical findings, due to the cascading compression, the training process has considerable deterioration on the convergence performance. To overcome this limitation, we implement a sign-bit compression-based learning synchronization framework, Marsit. It prevents cascading compression via an elaborate bit-wise operation for unbiased sign aggregation and its specific global compensation mechanism for mitigating compression deviation. The proposed framework retains the same theoretical convergence rate as non-compression mechanisms. Experimental results demonstrate that Marsit reduces up to 35\% training time while preserving the same accuracy as training without compression.


\end{abstract}






\maketitle

\blfootnote{$^\dagger$ Feijie Wu (harli.wu@connect.polyu.hk) and Shiqi He (shiqihe@cs.ubc.ca) contributed equally to this research.}
\blfootnote{$^*$ Song Guo (song.guo@polyu.edu.hk) and Zhihao Qu (quzhihao@hhu.edu.cn) are corresponding authors}
\vspace{-20px}

\section{Introduction}


In an era of data explosion, there is an increasing demand for various fields to launch AI-driven applications in image classification \cite{perez2020object}, natural language processing (NLP) \cite{roy2021application}, and so forth. Behind these applications are numerous models that have been fit in huge-size datasets such as ImageNet \cite{imagenet}. To minimize the development cost, cloud providers, e.g., Amazon AWS, offer various training paradigms to enable fast AI/ML solution deployment.


Nowadays, multi-hop all-reduce (MAR) training paradigm, including ring all-reduce (RAR) \cite{baidu, horovod} and 2D-torus all-reduce (TAR) \cite{mikami2018massively}, substitutes classical single-hop approaches such as parameter server (PS) and gossip, and becomes the most pervasive synchronization paradigm in high-performance computing (HPC) systems. 
For parallel stochastic gradient descent (PSGD) \cite{li2014scaling}, MAR achieves a better resource utilization under multi-GPU circumstance than PS.
Firstly, all GPUs involve in both the training and synchronization in MAR, while GPUs in PS architecture are categorized into two groups separately performing these two functionalities. Secondly, MAR prevents the network congestion at a single node because each client is not required to simultaneously process tremendous transmission requests. As a paradigm that workers are solely permitted to communicate with their neighbors, gossip has made great success in recent years \cite{lu2021optimal, lin2021quasi}. However, the performance of gossip in terms of convergence rate is much slower than MAR, especially under sparse connections such as ring topology \cite{chen2021accelerating}. 



\begin{table}[t]
\centering
\renewcommand{\arraystretch}{0.95}
  \begin{tabular}{lclcc}
  \Xhline{3\arrayrulewidth}
   & \multicolumn{2}{c}{Rounds} & Accuracy (\%) & Time (min)  \\\hline
  \multicolumn{5}{l}{cascading compression}\\
  \quad \quad $M=3$ & \makecell{\progressbar[width=1cm, ticksheight=0, linecolor=black, filledcolor=black]{0.52}} & 187 & 87.2 $\pm$ 2.31 & 11.2 \\ 
  \quad \quad $M=8$ & \makecell{\progressbar[width=1cm, ticksheight=0, linecolor=black, filledcolor=black]{1}} & 1K+ & divergence & NA \\ \hline
  \multicolumn{5}{l}{no compression}\\
  \quad \quad $M=3$ & \makecell{\progressbar[width=1cm, ticksheight=0, linecolor=black, filledcolor=black]{0.40}} & 129 & 99.1 $\pm$ 0.13 & 20.7 \\
  \quad \quad $M=8$ & \makecell{\progressbar[width=1cm, ticksheight=0, linecolor=black, filledcolor=black]{0.26}} & 76 & 99.2 $\pm$ 0.07 & 10.6 \\\Xhline{3\arrayrulewidth}
  \end{tabular}
  \caption{Training MNIST over AlexNet. The results show the best test accuracy by setting the stepsize in \{0.03, 0.01, 0.005\}.}
  \label{table:pretest}
  \vspace{-25px}
\end{table}

In network-intensive HPC systems such as public clouds, it is challenging to transfer a non-compressed gradient among nodes due to overwhelming bandwidth consumption. With the increasing size of a deep learning model, e.g., 60.2M weights on ResNet-152 \cite{resnet} and 100T on GPT-4 \cite{brown2020language}, the problem becomes severe because data transmission takes a significant amount of time. As a promising communication compression approach, signSGD \cite{bernstein2018signsgd_majority, safaryan2021stochastic, liu2018signsgd, tang20211} solely uses an element's sign to represent itself, where the number of encoding bits for each real number is dramatically deducted, i.e., from single float precision (32 bits) to 1 bit. 

Existing signSGD algorithms, albeit well-performed under PS, have limited performance under MAR, especially when the model is sufficiently large. Without a centralized coordinator to which each node independently sends its data, information asymmetry occurs when MAR leverages a sign matrix that includes all clients' gradients through cascading compression.
Each client inevitably performs decompression and then compression operation for transmission, accumulating errors. Although cascading compression can converge at the end for a small-scale environment, empirical studies in Table \ref{table:pretest} manifest its poor performance in comparison with the non-compressed algorithms. Also, more workers achieves better performance in non-compressed PSGD, whereas leading to divergence in the cascading compression scheme. 

To alleviate information asymmetry, we propose a framework for \underline{m}ulti-hop \underline{a}ll-\underline{r}educe using \underline{s}ign-b\underline{it}, named as Marsit. The core idea is to achieve unbiased sign aggregation by means of an elaborate bit-wise operation: The sign of an element remains unchanged if and only if it has the same sign in both vectors, while it follows a predefined probability distribution if it has different binary values. Such an operation supports that the reception and compression processes can take place in parallel. 
Furthermore, we introduce a global compensation mechanism to bridge the gap of compression error. The design is to equalize the clients' contribution towards final gradient because data on the cloud can be shuffled and formed an identical distribution among workers. 
To get rid of excess error accumulation, we periodically operate a full-precision transmission. 
Our contributions are summarized as follows: 
\vspace{-5px}
\begin{itemize} [leftmargin=.1in]
    \item Based on the designed one-bit operator and the global compensation scheme, we implement Marsit to support one-bit transmission without cascading compression under MAR. 
    \item We prove that the convergence rate for non-convex objectives is $O(1/\sqrt{TM})$ under RAR framework, where $T$ and $M$ represent the numbers of synchronizations and workers, respectively. The theoretical result indicates that our algorithm achieves a linear speedup simultaneously with respect to the number of workers. To the best of our knowledge, this is the first work that addresses information asymmetry under MAR; 
    \item We conduct an empirical study to illustrate the effect of our proposed algorithms on RAR and TAR. It is conducted with ResNet-50 on ImageNet for image recognition. It reduces the communication cost by around 90\% as compared with non-compressed methods while preserving the same convergence performance. 
\end{itemize}


\vspace{-10px}
\section{Related Works} \label{sec:related_work}

\paragraph{Quantization} At the cost of the gradient precision, quantization approaches reduce the number of encoding bits for each real number \cite{qsgd_tern_wen, qsgd, qsgd_zip_zhang}. Although GradiVeQ \cite{yu2018gradiveq} utilizes singular value decomposition to achieve linear quantization under RAR, the process requires considerable computation consumption such that the receiving period cannot cover the time length of compression. 
\vspace{-5px}
\paragraph{signSGD} As a promising approach, signSGD represents the elements of a gradient using their signs, which reduces the communication overhead by 32$\times$ at every iteration \cite{bernstein2018signsgd_theorem}. It has remarkable performance under PS, including 1-bit Adam \cite{tang20211}, SSDM \cite{safaryan2021stochastic} and majority vote \cite{bernstein2018signsgd_majority}. However, they are not suitable for MAR since their aggregation process cannot guarantee within one bit at each transmission. 
\vspace{-5px}
\paragraph{Other communication compression approaches} There are various approaches to reduce communication overhead, such as sparsification \cite{wangni2018gradient, guo2020tail} and low rank \cite{vogels2019powersgd}. However, these approaches may not have well performance for MAR under some network topologies. For instance, PowerSGD \cite{vogels2019powersgd} requires to transmit multiple sequential vectors at a synchronization, which undermines the training efficiency under RAR. 

\vspace{-5px}
\section{Motivation} \label{sec:preliminary}

\textbf{Objectives.} Under an $M$-worker MAR, the objective is to minimize the cumulative expected loss, which can be formulated as 
\vspace{-1px}
\begin{equation} \label{equation:problem}
    \min_{\bm{x} \in \mathbb{R}^d} \quad F(\bm{x}) = \frac{1}{M} \sum_{m=1}^{M} \underbrace{\mathbb{E}_{\xi_m \sim \mathcal{D}_m} \left[f_m(\bm{x}, \xi_m)\right]}_{:= F_m(\bm{x})},
\end{equation}
where $\mathcal{D}_m$ is the local data distribution on worker $m$, $f_m(\bm{x}, \xi_m)$ is the empirical loss given parameter $\bm{x}$ and stochastic sample $\xi_m$ from $\mathcal{D}_m$, and $F_m(\cdot)$ is an objective function. Given that the entire training locates in the cloud, we assume that all the local datasets have an equal size. Since the objective function is randomly extracted over a given data distribution, it is a common practice that the bias does not exist between the expected loss and the empirical one. 


\subsection{Why Bit Length Expansion Occurs?}

\begin{figure}[b]
\centering
\vspace*{-20pt}
    \subfloat[Time length per iteration]{
    \centering
        \includegraphics[width=0.31\textwidth]{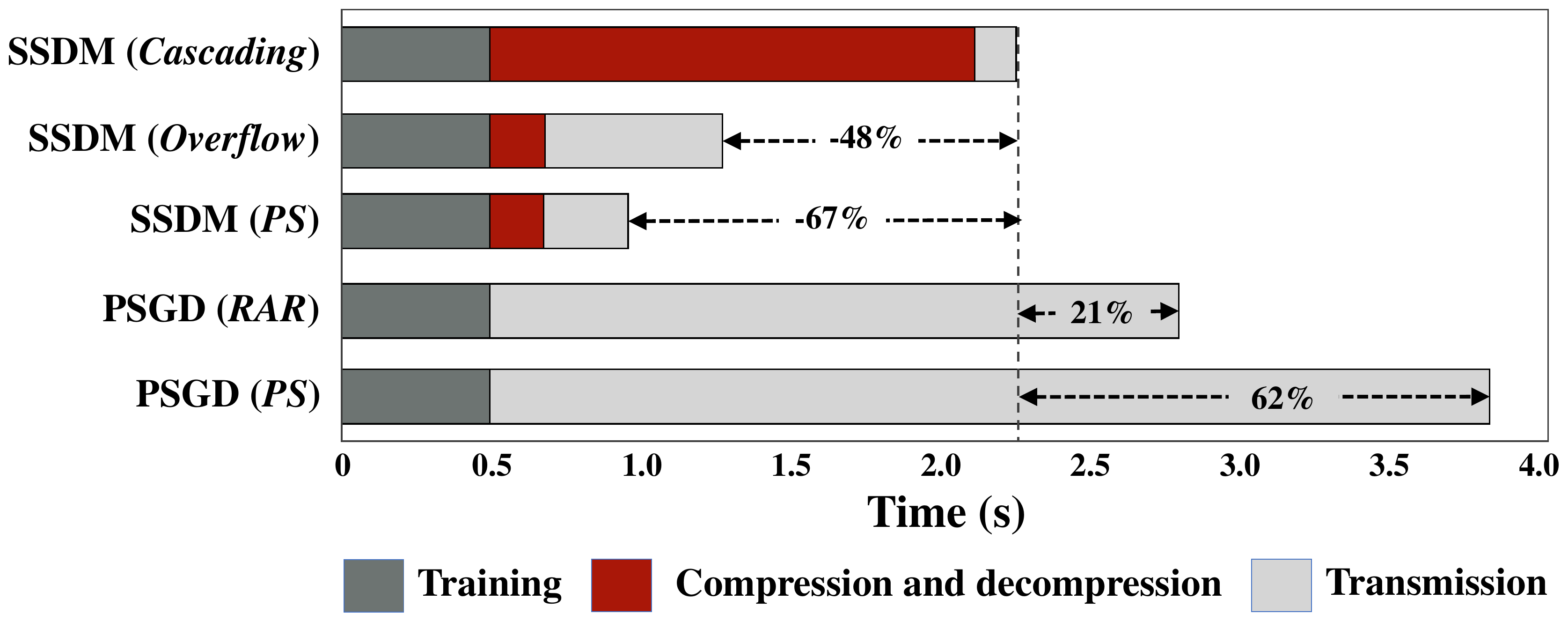}
        \label{fig:pretest_time}
    }
    \subfloat[Matching rate]{
    \centering
        \raisebox{.0\height}{\includegraphics[width=0.17\textwidth]{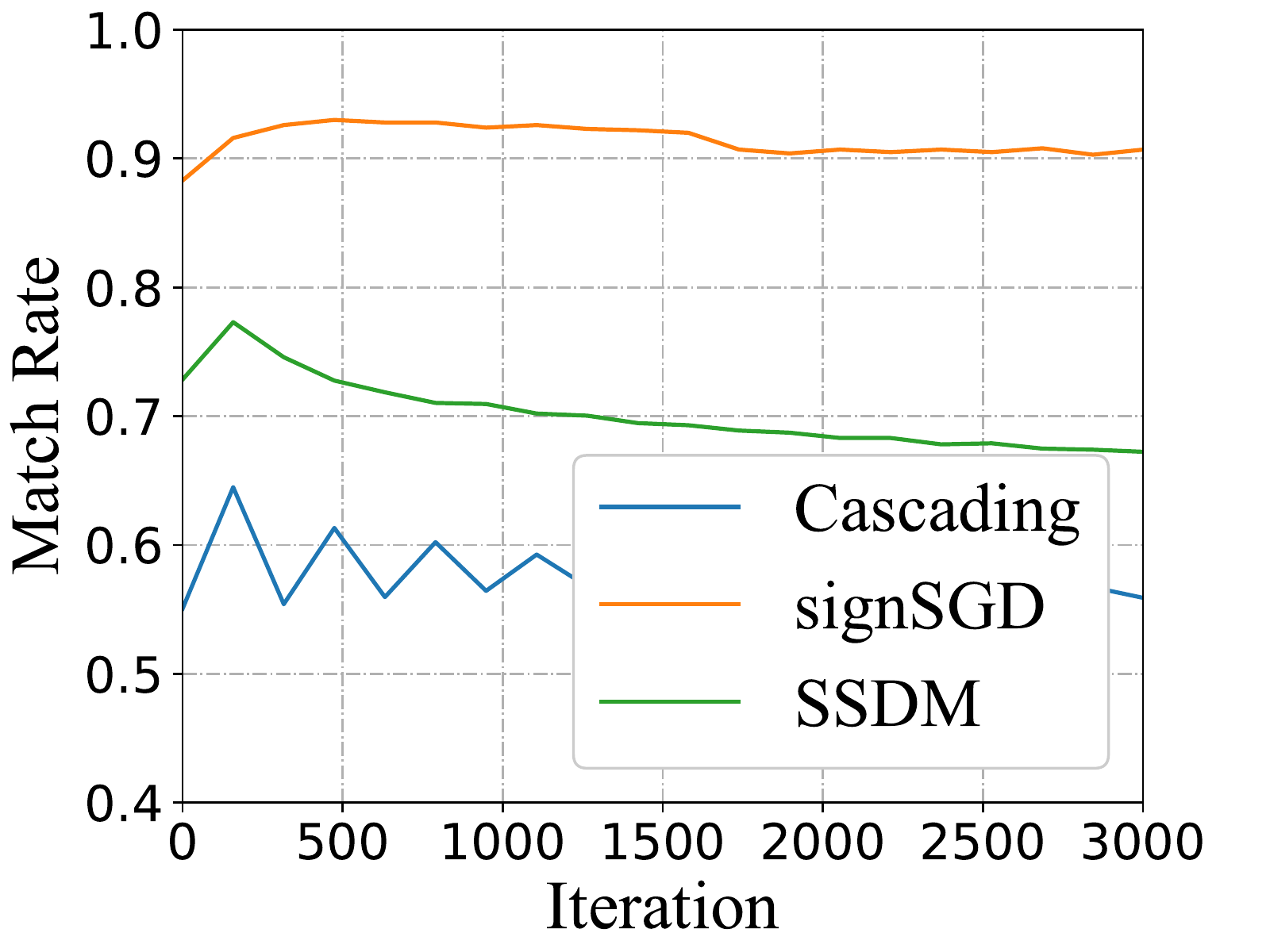}}
        \label{fig:mismatching}
    }
    \vspace{-10pt}
    \caption{Training MNIST over AlexNet with 3 workers. The comparison of existing approaches on an iteration's training time length and matching rate. }
    \vspace{-5pt}
    \label{fig:time_proportion}
\end{figure}

In a non-compressed algorithm, MAR naturally requires less communication overhead than PS when synchronizing a model among all nodes. For example, given a $D$-dimension neural network, RAR requires the consumption of $2 \times (M-1) \times D$ weights, while PS needs that of $2 \times M \times D$. In Figure \ref{fig:pretest_time}, non-compressed approach under RAR costs less time than the one under PS. 

An operation compatible to MAR should be linear, which allows workers directly aggregate without extra decompression-compression process \cite{vogels2019powersgd}. SSDM \cite{safaryan2021stochastic} is one of rare signSGD approaches that satisfy the requirement of linearity, where its aggregation is to sum up all the sign bit. With the operation, workers do not fit the transmission elements into one bit under MAR synchronization, but with an upper bound of $\left\lceil\log_2 M\right\rceil$. As shown in Figure \ref{fig:pretest_time}, such way spends longer time than its PS solution in transmission period due to the growing size of transmission packages. Therefore, the approach is not efficient under MAR settings and we are dedicated to implementing a compression framework that restricts the transmission size by only one bit. 

\begin{figure*}
    \centering
    \includegraphics[width=1\textwidth]{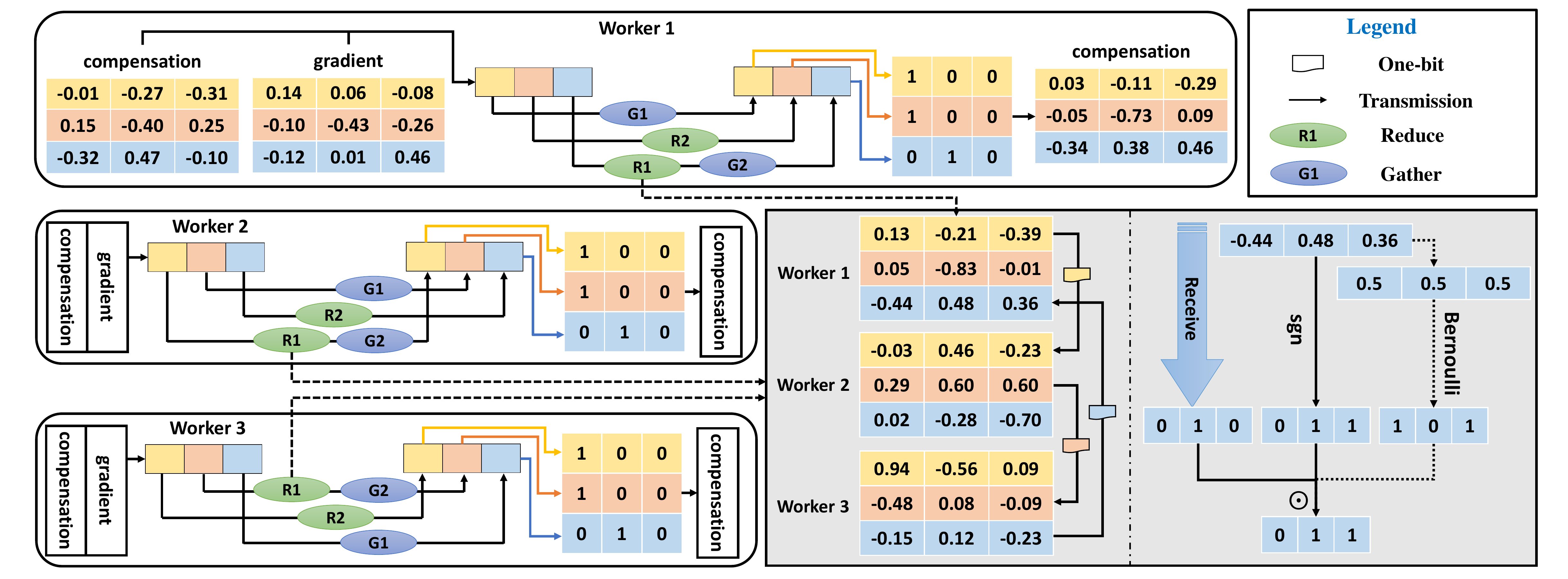}
    \vspace{-20pt}
    \caption{The workflow of Marsit under ring network topology with a total of three workers}
    \vspace{-10pt}
    \label{fig:pipeline}
\end{figure*}

\subsection{Why Not Cascading Compression?} 

For a $D$-dimension vector $\bm{g}$, SSDM \cite{safaryan2021stochastic} (denote by $\mathcal{Q}$) compresses an element $g_i$ ($i \in \{0, ..., D-1\}$) consistent with its sign following the probability of $\frac{1}{2} + \frac{|g_i|}{\|\bm{g}\|_2}$, where $\|\cdot\|$ means $\ell_2$-norm. Apparently, it is an unbiased compression method. To ensure each transmission limited in one bit, a client performs the step-by-step sequence: 
\begin{itemize}[leftmargin=.2in]
    \item \textbf{Receive} aggregated gradient segment(s), including corresponding $\ell_2$-norm(s) and sign vector(s), from the last worker(s); 
    \item \textbf{Recover} the gradient segment(s) as $\bm{w}$ for full precision;
    \item \textbf{Aggregate} local gradient $\bm{v}$ with decompressed segment(s); 
    \item \textbf{Compress} the assembled segment into a precision-loss one, i.e., $\mathcal{Q}(\bm{w} + \bm{v})$;
    \item \textbf{Send} the compressed segment to the next worker(s). 
\end{itemize}
The workflow, named as cascading compression, is able to broadcast and unify the updates among clients. Obviously, the expected result of cascading compression is equivalent to the sum of all gradients.
However, cascading compression has two major shortcomings.

\subsubsection{Overwhelming Time Consumption} 
In contrast to \textit{all-reduce} operation, cascading compression inevitably requires additional decompression-compression steps, which spends more time on synchronization. In Figure \ref{fig:pretest_time}, although the approach has significant improvement in terms of communication, the decompression-compression period consumes a large amount of time, making it inefficient when comparing with the one uder PS and under all-reduce operation. 


\subsubsection{Performance Deterioration}
Notably, the second step cannot actually represent the real aggregation results. In this case, the error accumulates and spreads over the network, which deteriorates the training performance. Besides, it is not suitable to use the $\ell_2$-norm to achieve unbiased compression because its value is so large that the new compressed sign is more likely biased to the received one, even if the actual aggregation sign should be the opposite one. As demonstrated in Figure \ref{fig:mismatching}, among the applicable settings, cascading compression has the lowest matching rate (i.e., around 56\%) measured by the sign of non-compression aggregation value. Following remark compares the performance between cascading compression under RAR and centralized training under PS. 

\noindent\textbf{Remark.} We assume that the $\ell_2$-norm of any gradients are bounded by a non-negative scalar $G$. Suppose SSDM \cite{safaryan2021stochastic} is achieved as unbiased estimator under centralized training and cascading compression, where the expected update value is equivalent to the update of non-compression algorithm. For training a deep neural network where the value of $D$ is quite large, the upper bound of gradient deviation, i.e., the Euclidean distance between the expected result and the actual update, for cascading compression explodes rapidly with $M$, while centralized training does not exist.\footnote{The detailed proofs for the remark and Theorem \ref{theo:marsit} in this paper is available in the appendix} 

\section{Marsit} \label{sec:algorithm}

In this section, we first provide a holistic insight for Marsit. Then, in Section \ref{subsec:details} and Section \ref{subsec:theory}, we present the technical details and the theoretical analysis, respectively. 

Due to the lack of centralized server under MAR, all workers should maintain a global model locally, the parameters of which are always consistent with others. Figure \ref{fig:pipeline} illustrates the pipeline of Marsit under RAR, a common paradigm for MAR using ring network topology.
Each worker possesses a compensation vector and a gradient, and aggregates them into a standalone vector. Then, they partition the vector into several segments and exchanges them at the synchronization phase which consists of a reduce period (highlighted in green and marked as R) and a gather period (highlighted in blue and marked as G). In the reduce period, i.e., R1 and R2, a worker processes the received message with corresponding local segment and sends it to the next worker. Here we exemplify with R1 and depict the procedures in the gray box. The left part of the gray box presents how the message transfers among workers, while the right takes the message transferring from worker 3 to worker 1 (highlighted in azure) as an example and exhibits how to aggregate the received vector and the local vector. In the gather period, i.e., G1 and G2, a worker substitutes corresponding local segment with the received information and transmits it to the next worker. The relevant processes have been widely adopted in \cite{baidu, horovod}. After the synchronization phase, all clients reach to a consensus and holds the same gradient which is used to update the global model and the local compensation vector.

\subsection{Implementation Details} \label{subsec:details}

Here we discuss the key operations with in-depth justifications. Generally, the workflow lies in two phases: one-bit synchronization in each round to reduce the communication cost, and full-precision synchronization executed every $K$ rounds to periodically eliminate the error accumulation. The full implementation is given in Algorithm \ref{algo:marsit} to demonstrate the workflow behind Marsit, and Algorithm \ref{algo:marsit_sgd} is to illustrate how we can apply Marsit to existing optimizers like stochastic gradient descent (SGD). 



\subsubsection{Global Model Synchronization (Line 4--8 in Algorithm \ref{algo:marsit})}
No matter which phase it is, Marsit synchronizes the gradients through MAR. Full-precision synchronization has been widely discussed in the previous studies \cite{baidu, horovod, jia2018highly, mikami2018massively}, which is equivalent to the aggregation result under PS, we mainly focus on the synchronization using sign bit only in this part. 

As illustrated in Figure \ref{fig:pipeline}, both receiving vector $v_i$ and local compression $v_i^*$ (Line 5 in Algorithm \ref{algo:marsit}) run in parallel, which reduces a great amount of time in comparison with the cascading compression. Since both $v_i$ and $v_i^*$ are a sign-bit vector, a problem raises on how to aggregate both vectors without additional compression-decompression processes. Therefore, we define a novel bit-wise operator $\odot$ to ensure these two vectors compatible with each other. In this update process, if an index on both vectors is the same, then the transmission vector at this points remains unchanged. However, considering element inconsistency between $v_i$ and $v_i^*$, we use a transient vector, $v$, which predetermines the transmitted binary value when confronted with inconsistent elements. It follows a Bernoulli distribution: Let $b_j$ be the probability for the element $j$ of vector $v_i^*$ (denote by $v_{i, j}^*$) at worker $m$ that marks as 1 in vector $v$: 
\begin{equation}
    b_j = 
        \begin{cases}
            (m-1)/m & v_{i, j}^* = 0 \\
            1/m & v_{i, j}^* = 1 \\
        \end{cases} \overset{\text{Bernoulli}}{\implies} v_j = 
        \begin{cases}
            1 & pr=b_j \\
            0 & \text{Otherwise} \\
        \end{cases}
\end{equation}
Note that the process can take place in parallel with the receiving stage but after the calculation of $v_i^*$. With the transient vector $v$, the updated operator $\odot$ between $v_i$ and $v_i^*$ should be expressed as: $v_i \odot v_i^* = (v_i \text{ AND } v_i^*) \text{ OR } (v_i \text{ XOR } v_i^* \text{ AND } v)$. By mathematical analysis, the expected value of the sign bit is equivalent to the average of the sign bits among all clients. 

\setlength{\textfloatsep}{0.3cm}
\begin{algorithm}[t]
\setstretch{0.5}
\SetKwData{Left}{left}\SetKwData{This}{this}\SetKwData{Up}{up}
\SetKwFunction{Union}{Union}\SetKwFunction{FindCompress}{FindCompress}
\SetKwInOut{require}{Require}\SetKwInOut{return}{Return}
\require{Synchronization index $t$, number of communication rounds for full-precision synchronization $K$, gradient $g_t^{(m)}$, compensation vector $c_t^{(m)}$, global stepsize $\eta_s$}
Calculate the update by $g_t^{(m)} \leftarrow g_t^{(m)} + c_t^{(m)}$\;
Split $g_n^{(m)}$ into $M$ parts, and denote by {$g_{t, i}^{(m)}, \forall i \in \{0, ..., M-1\}$}\;
\uIf{\text{mod}(t, K) $\neq$ 0} {
\For{$  i\leftarrow 0$ \KwTo $M-1$}{
Receive the sign vector $v_i$ in parallel with\\
\nonl \quad $\bullet$ Calculate the sign vector by $v_i^{*} \leftarrow \text{sgn}\left(g_{t,i}^{(m)}\right)$\;
Update the transmission sign vector via $v_i \leftarrow v_i \odot v_i^*$\;
Send $v_i$ to the next worker\;
}
Aggregate the global update via $g_t \leftarrow \eta_s \cdot \left(\bigcup_{i=0}^{M-1} v_i\right)$\;
Update compensation vector via $c_{t+1}^{(m)} \leftarrow g_t^{(m)} - g_t$\;
}
\Else {
Aggregate the global update via $g_t \leftarrow \frac{1}{M} \sum_{m=1}^M g_t^{(m)}$\;
Update compensation vector via  $c_{t+1}^{(m)} \leftarrow 0$\;
}
\return{The global update $g_t$, compensation vector $c_{t+1}^{(m)}$}
\caption{Marsit (worker $m$)}\label{algo:marsit}
\end{algorithm}
\setlength{\floatsep}{0.1cm}

\vspace{-3px}



\subsubsection{Global Model Update (Line 9 and Line 12 in Algorithm \ref{algo:marsit})}
The value $g_t$ depends on whether the synchronization is under full precision. If the transmission is sign-bit only, Line 9 returns $g_t$ that comes from a vector of signs multiplying a global learning rate. 
As for full-precision synchronization in Line 11, extra learning rate is not necessary since $g_t^{(m)}$ has included the local stepsize. The purpose for this update is to eliminate the accumulated error and accelerate the training process. In Figure \ref{fig:various_K}, we demonstrate there exists a trade-off between the final accuracy and the additional communication costs due to full precision synchronizations, by choosing different system parameter $K$. 

\vspace{-5px}
\subsubsection{Global Compensation Mechanism (Line 10 and Line 13 in Algorithm \ref{algo:marsit})} 
\begin{figure}[t]
    \captionsetup[subfloat]{farskip=2pt,captionskip=1pt}
    \centering
    \subfloat[Epoch and Accuracy]{
        \raisebox{-.43\height}{\includegraphics[width=.24\textwidth]{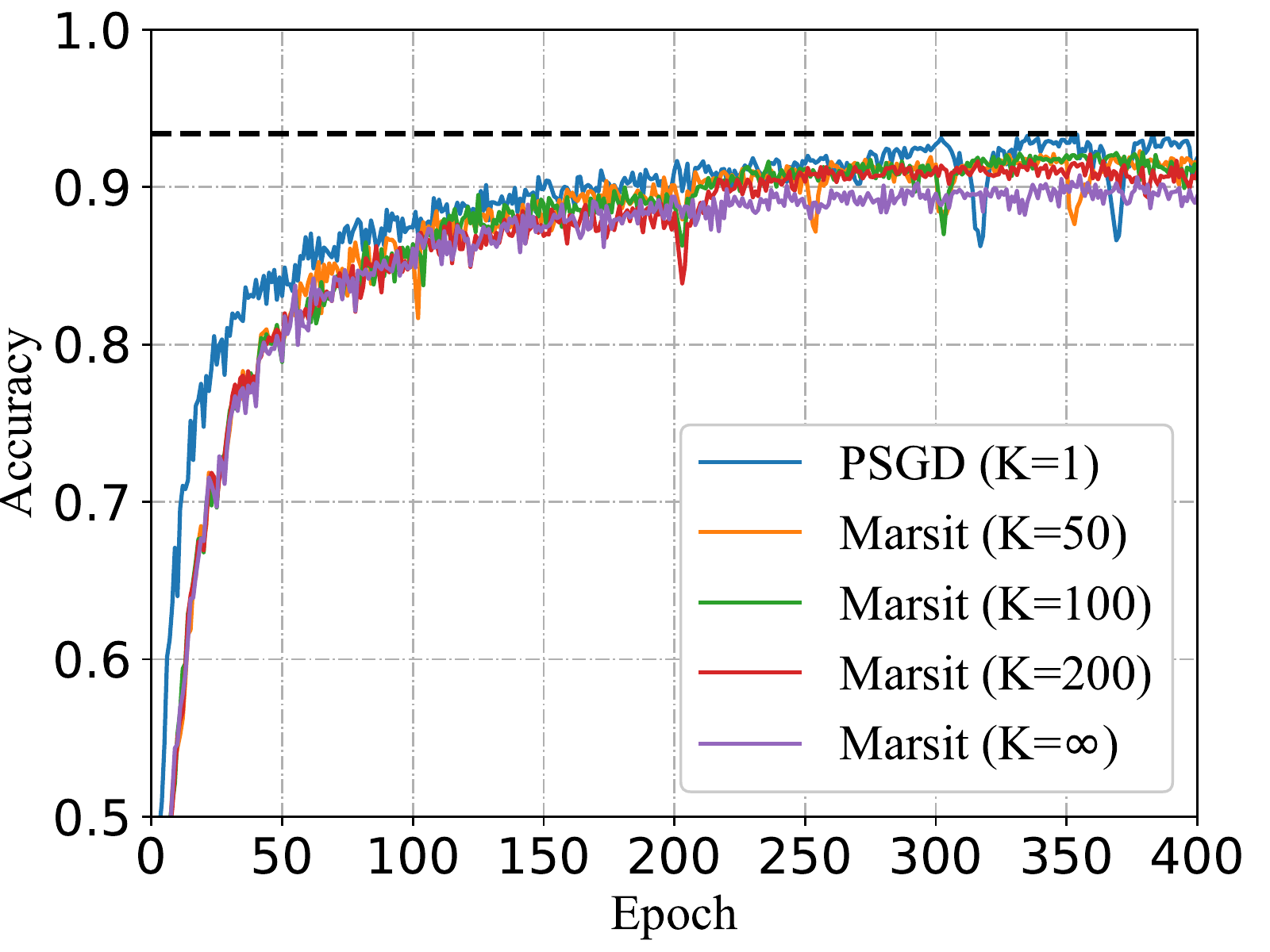}}
        \label{fig:g_value}
    }
    \subfloat[Convergence results]{
        \vspace*{100px}
        \small
        \begin{tabular}{cccc}
        \hline
            $K$ & \begin{tabular}[c]{@{}c@{}}Time \\ (min)\end{tabular} & \begin{tabular}[c]{@{}c@{}}Acc.\\ (\%)\end{tabular} & Bits \\ \hline
            1 & 40.18 & 93.42 & 32 \\
            50 &  22.05 & 92.28 & 1.62\\
            100 & 21.34 & 91.73 & 1.31 \\
            200 & 22.38 & 92.00 & 1.16 \\
            $\infty$ & 18.78 & 90.75 & 1  \\ \hline
            \\
        \end{tabular}
    }
    \vspace{-10px}
    \caption{Training CIFAR-10 over AlexNet by evaluating various values of $K$. (a) indicates the relation between epoch and accuracy; and (b) depicts the convergence result. $K = \infty$ means $K$ is greater than the maximum communication rounds, i.e., 400 in this case. }
    \label{fig:various_K}
    \vspace{-5px}
\end{figure}

At the beginning of the model training, we initialize the local compensation gradient with a zero vector (Line 1 in Algorithm \ref{algo:marsit_sgd}) by default. All clients have the consensus on how to update the global model, i.e., $g_t$ at Line 9 in Algorithm \ref{algo:marsit}, which is a vector containing binary value only to indicate the sign of each element. Unlike traditional compensation approaches under single-hop synchronization, a client in Marsit cannot obtain how much it contributes to the aggregation under multi-hop synchronization. 
Based on the independent and identical data distribution on cloud training, every client compresses and obtains the same gradient in expectation. Thus, we apply an identical local compensation amount for each client, which then combines into the global compensation.
Considering the accumulated error could be quite large, we periodically reset the error by means of full-precision synchronization, where the compensation vector can be set to 0. As we can see in Figure \ref{fig:various_K}, although greater $K$ costs less time to reach the stable point, they have smaller convergence accuracy. Also, greater $K$ may not always speed up the convergence progress, for instance, when $K$ changes from 100 to 200, more time is required to realize the convergence feature.

\begin{table*}[t] 
\centering
\begin{tabular}{|c|c|c|c|c||c|c|c|c|c|}
\hline
\multirow{2}{*}{Model}
& \multirow{2}{*}{Dataset}
& \multirow{2}{*}{\# parameters}
& \multirow{2}{*}{Batch size}
& \multicolumn{6}{c|}{Top-1 Accuracy (\%)}\\
\cline{5-10}
 & & & & PSGD & signSGD & EF-signSGD & SSDM  & Marsit-100 & Marsit

\\
\hline
 

 AlexNet & CIFAR-10 & 23M & 8192 & 82.38 & 80.74 & \textit{82.25} & 81.89 & \textbf{82.30} & 81.58  \\
 \hline
 ResNet-20 & CIFAR-10 & 0.27M & 8192 & 93.42 & 88.92 & \textit{91.85} & 89.18 & \textbf{92.18} & 90.15  \\
 \hline
 ResNet-18 & ImageNet & 11M & 6144 & 69.18 & 67.17 & 68.14  & 68.10 & \textbf{68.96} & \textit{68.40} \\
 \hline
 ResNet-50 & ImageNet & 25M & 6144 & 74.87 & 72.74 & 73.89 & 73.35 & \textbf{74.35} & \textit{74.10} \\
\hline
DistilBERT & IMDb review & 8.3B & 512 & 92.16 & 89.12 & \textit{90.57} & \textbf{91.41} & 90.13 & 90.26\\
\hline
\end{tabular}
\caption{Accuracy of existing works on different models training for different datasets.}
\label{table:2}
\vspace{-20pt}
\end{table*}

\subsection{Theoretical Guarantees} \label{subsec:theory}

\begin{algorithm}[!t]
\setstretch{0.8}
\SetKwData{Left}{left}\SetKwData{This}{this}\SetKwData{Up}{up}
\SetKwFunction{Marsit}{Marsit}\SetKwFunction{FindCompress}{FindCompress}
\SetKwInOut{Input}{Input}\SetKwInOut{Output}{Output}
\Input{Initial Point $\tilde{\bm{x}}_0$, local stepsize $\eta_l$, global stepsize $\eta_s$, number of communication rounds for full-precision synchronization $K$, number of global synchronizations $T$}
Initialize local compensation gradient $c_0^{(m)} \leftarrow \mathbf{0}$\;
\For{$  t\leftarrow 0$ \KwTo $T-1$}{
Randomly sample $\xi_k^{(m)}$ from local data $\mathcal{D}_m$\;
Compute local stochastic gradient $g_t^{(m)} \leftarrow \nabla f_m\left(\tilde{\bm{x}}_t; \xi_{k}^{(m)}\right)$\;
$g_t, c_{t+1}^{(m)} \leftarrow$ \Marsit{$t$, $K$, $\eta_l g_t^{(m)}$, $c_{t}^{(m)}$, $\eta_s$}\;
Update the parameters through $\tilde{\bm{x}}_{t+1} \leftarrow \tilde{\bm{x}}_{t} - g_t$\;
}
\Output{The final model $\tilde{\bm{x}}_T$}
\caption{Marsit-driven SGD (worker $m$)}\label{algo:marsit_sgd}
\end{algorithm}

To theoretically analyze the convergence results for Marsit, we have the following assumption for Problem (\ref{equation:problem}), which are ubiquitously applied to \cite{bernstein2018signsgd_theorem, safaryan2021stochastic, guo2020tail}. 
\begin{assumption} \label{assumption:general}
Problem (\ref{equation:problem}) satisfies the following constraints: 
\begin{enumerate}[leftmargin=.15in]
    \item \textbf{Smoothness:} All function $F_m(\cdot)$'s are continuous differentiable and their gradient functions are $L$-Lipschitz continuous with $L>0$; 
    \item \textbf{Bounded variance:} For any worker $m$ and vector $\bm{x} \in \mathbb{R}^d$, there exists a scalar  $\sigma \geq 0$ such that {\footnotesize $\mathbb{E}_{\xi \sim \mathcal{D}_m} \|\nabla f_m(\bm{x}, \xi) - \nabla F_m (\bm{x})\|^2_2 \leq \sigma^2$}. 
\end{enumerate}
\end{assumption}
\noindent Based on the preceding assumptions, the following theorem holds: 
\begin{theorem} \label{theo:marsit}
Under Assumption \ref{assumption:general}, by setting local learning rate for $\eta_l = \sqrt{M/T}$ and the global learning rate $\eta_s = \sqrt{1/TD}$, the upper bound for Algorithm \ref{algo:marsit_sgd} using RAR-based  should be:
\begin{equation*}
    \min_{t \in \{0, ..., T-1\}}\mathbb{E} \left\| \nabla F(\tilde{\bm{x}}_{t}) \right\|_2^2 \leq \mathcal{O}\left(\frac{1}{\sqrt{MT}}\right) + \mathcal{O} \left(\frac{K(K+1)}{T}\right)
\end{equation*}
where we treat $F_*-F(\tilde{\bm{x}}_{1})$, $L$ and $\sigma$ as constants. 
\end{theorem}
\noindent \textbf{Remark.} Given that the value of $K$ is much smaller than the value of $T$, our approach can achieve a convergence rate of $O(1/\sqrt{MT})$, which achieves linear speedup with the number of the workers. In other words, the more GPUs participate in the model training, the faster Marsit reaches a stable point. 




\section{Experimental Setup} \label{sec:setup}

We evaluate our proposed framework on scenarios that meet the requirement of current industrial needs and cover the most representative model training instances on the public clouds. In this section, the problem we explore mainly lies in these two categories: (i) whether there exists a significant accuracy drop in comparison with non-compression methods; (ii) how fast a model achieves convergence in comparison with existing compression approaches under MAR. 

\paragraph{Datasets, models and tasks} Our experiments consist of three datasets: CIFAR-10 \cite{cifar10}, ImageNet \cite{imagenet} and IMDb reviews \cite{maas-EtAl:2011:ACL-HLT2011}. The first two datasets are frequently used for image classification and consist of 60K 32$\times$32 and 14M 224$\times$224 colored images, respectively. The last one is for sentiment analysis with 50K movie reviews. The models vary among the datasets: AlexNet \cite{krizhevsky2012imagenet} and ResNet-20 \cite{resnet} for CIFAR-10, ResNet-18 and ResNet-50 for ImageNet, and DistilBERT \cite{sanh2019distilbert} for IMDb reviews. 

\paragraph{Implementation}
The experiments are conducted on Huawei Cloud, where we deploy a cluster with 32 nodes and each node carries 2 Nvidia T4 GPUs. The underlying training framework is supported by Pytorch distributed computing package\footnote{\url{https://pytorch.org/tutorials/intermediate/dist_tuto.html}}. 
We implement Marsit on RAR \cite{baidu, horovod}, a classical MAR implementation over ring network topology, and 2D-torus all-reduce (TAR), a state-of-the-art MAR scheme over 2D-torus network topology. Marsit can be easily extended to other all-reduce paradigms including segmented-ring all-reduce \cite{jia2018highly} and tree all-reduce \cite{vogels2019powersgd}. 


\paragraph{Baselines} We implement multiple baselines to evaluate the performance of Marsit. PSGD \cite{li2014scaling} is implemented under MAR with full precision, i.e., 32 bits. For EF-signSGD \cite{karimireddy2019error}, signSGD with majority vote \cite{bernstein2018signsgd_theorem} and SSDM \cite{safaryan2021stochastic}, we extend them to MAR by dynamically changing the bit length. We also utilize Elias coding \cite{elias1975universal} to compact the transmission message among nodes. 

\paragraph{Optimizers and hyper-parameters} To reduce the frequency of the communications among nodes, clients perform multiple local updates between two successive synchronizations. The optimizer for image classification task is Momentum, and Adam for sentiment analysis. Marsit-100 refers to the setting where local gradients operate full-precision synchronization every 100 communication rounds (i.e., $K = 100$), while Marsit does not have full-precision synchronization. For ImageNet and CIFAR-10, the initial learning rate is set to 0.1 and 0.03, respectively, and decays by a factor of 10 every full-precision synchronization. For DistilBERT, we use a constant learning rate of 5e-5. 


\section{Numerical Results and Analysis} \label{sec:experiments}



\paragraph{Performance Analysis}
Table \ref{table:2} summarizes Top-1 accuracy of all test datasets. Compared to PSGD, the state-of-the-art compression approaches suffer from a noticeable accuracy drop in both image classification and sentiment analysis tasks. For instance, signSGD has up to 5\% decreasing. 
Moreover, in most cases, Marsit-100 and/or Marsit outperforms the existing approaches and achieves nearly the same final accuracy as PSGD. In CIFAR-10 training, Marsit with periodical full-precision synchronization (e.g., Marsit-100) has better performance than the one without full-precision synchronization, while they do not have distinct differences in both ImageNet and IMDb review datasets.

\begin{figure}[!t]
\centering
\vspace{-15px}
\subfloat[Accuracy w.r.t Time]{
\centering
\includegraphics[width=0.24\textwidth]{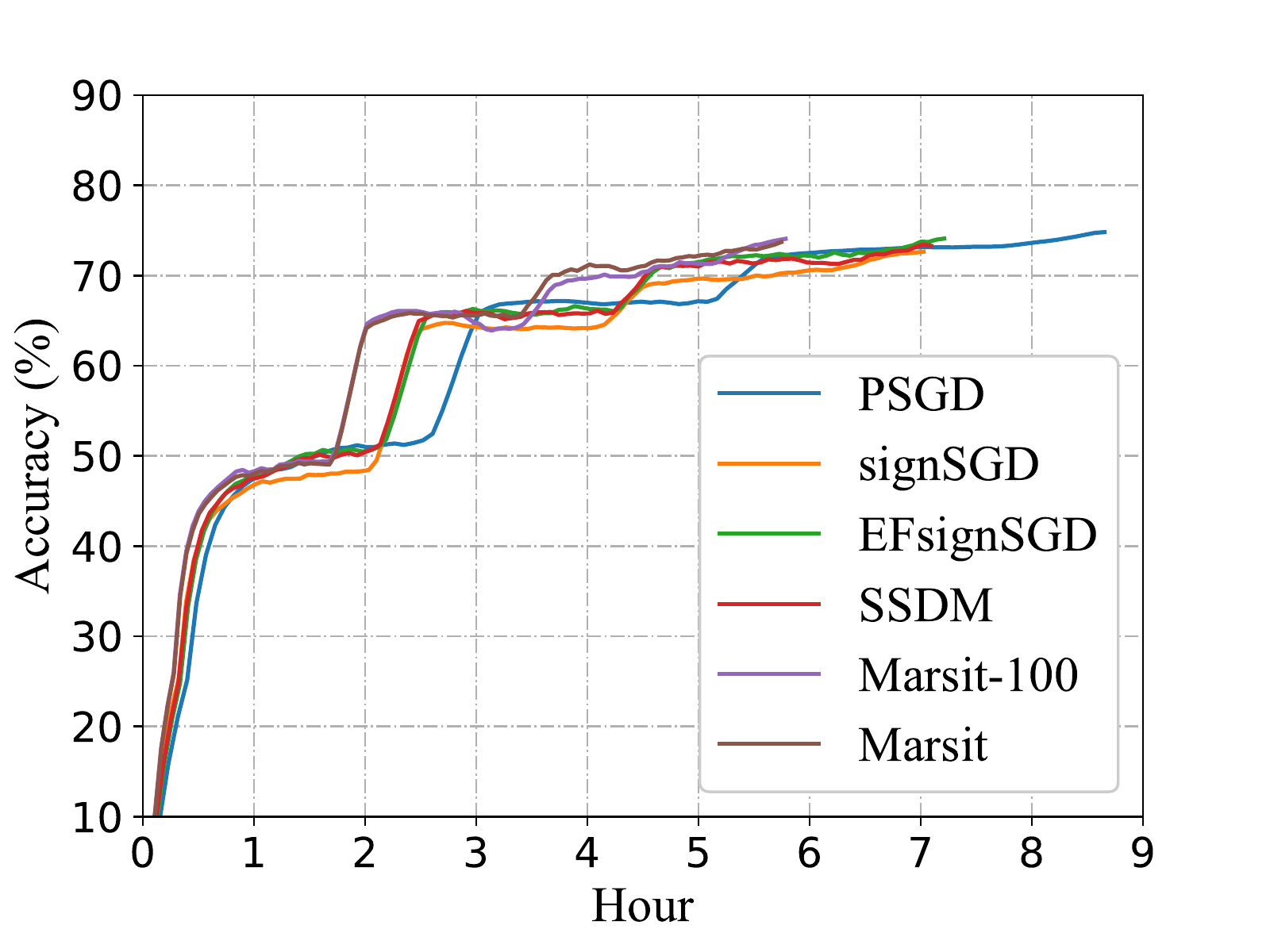}
\label{fig:resnet_time}
}%
\subfloat[Accuracy w.r.t Overhead]{
\centering
\includegraphics[width=0.24\textwidth]{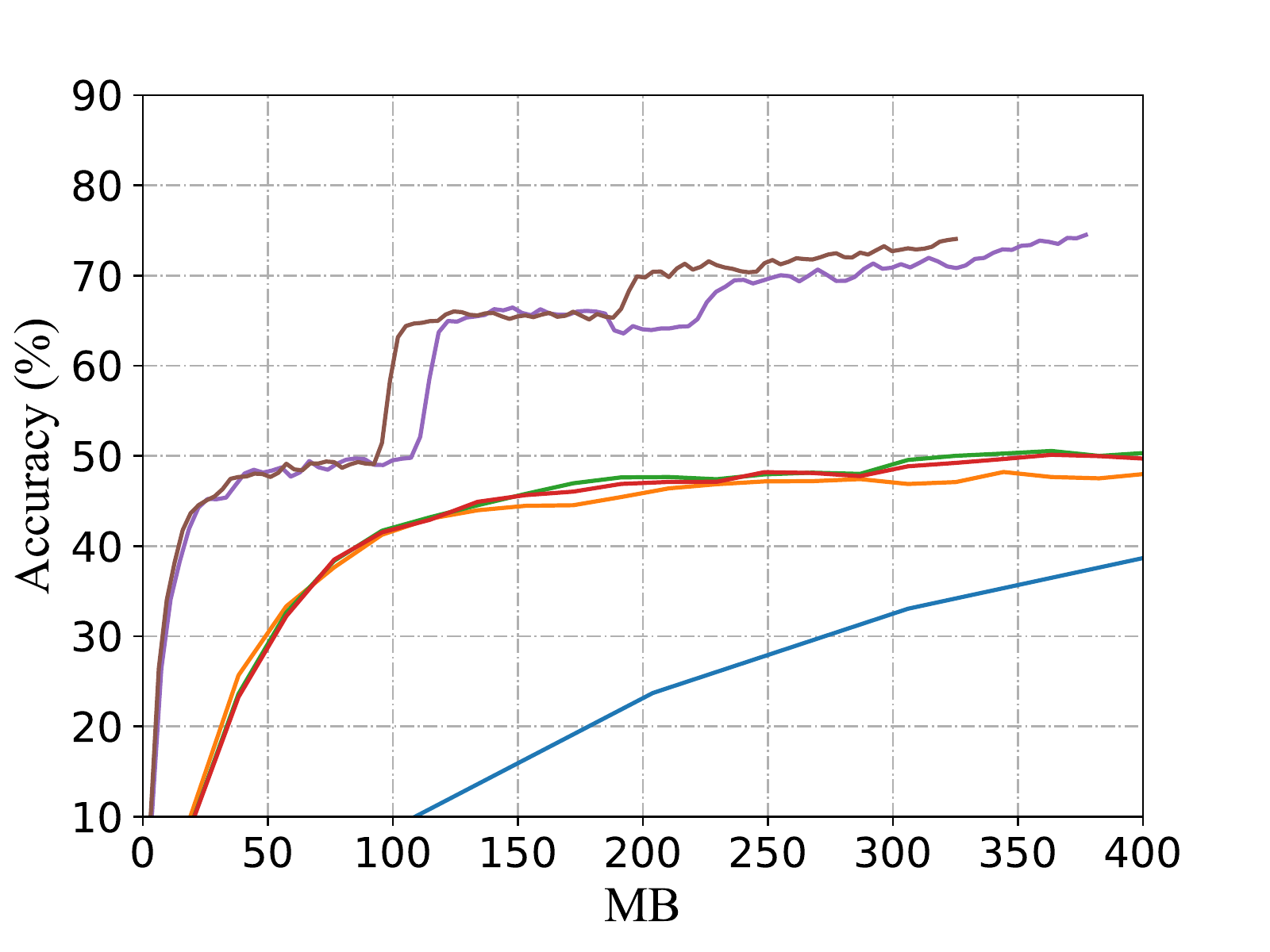}
\label{fig:resnet_bandwidth}
}%
\centering
\vspace{-10pt}
\caption{Experiments for training ResNet50 on ImageNet}
\label{fig:exp_general_case}
\end{figure}

Figure \ref{fig:resnet_time} shows time-to-accuracy performance for ResNet-50 on ImageNet. Among these six approaches, non-compression approach, i.e., PSGD, takes a large amount of time, while Marsit achieves large speedups (1.5x) to reach a similar accuracy. 

\paragraph{Communication Efficiency}  Marsit has a significant reduction in communication cost compared to the other five baselines. From Figure \ref{fig:resnet_bandwidth}, our algorithm requires 90\% less communication budget, when compared to PSGD, and reduces communication cost by 70\%, when compared to the existing signSGD approaches. In the mean time, with a smaller communication budget, our algorithm still preserves the same convergence rate as other baselines.

In Figure \ref{fig:resnet_bandwidth}, given the same amount of communication overhead, Marsit and Marsit-100 always have higher accuracy than other baselines. Specifically, when Marsit and Marsit-100 reach convergence, other signSGD methods only attain accuracy around 50\%.


\begin{figure}[!t]
\centering
\vspace{-10px}
\subfloat[TAR]{    
\centering
\includegraphics[width=0.235\textwidth]{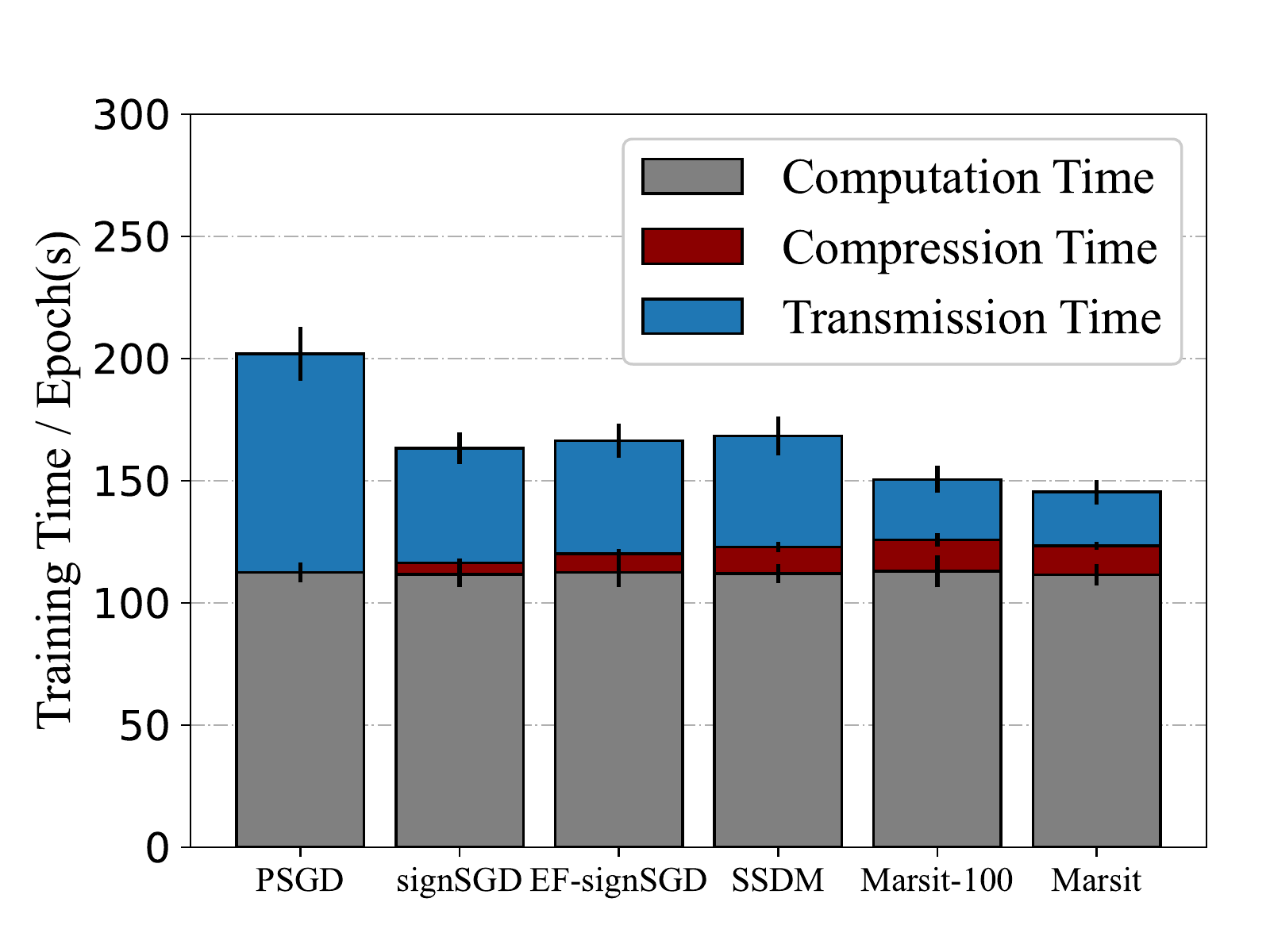}
}%
\subfloat[RAR]{
\centering
\includegraphics[width=0.235\textwidth]{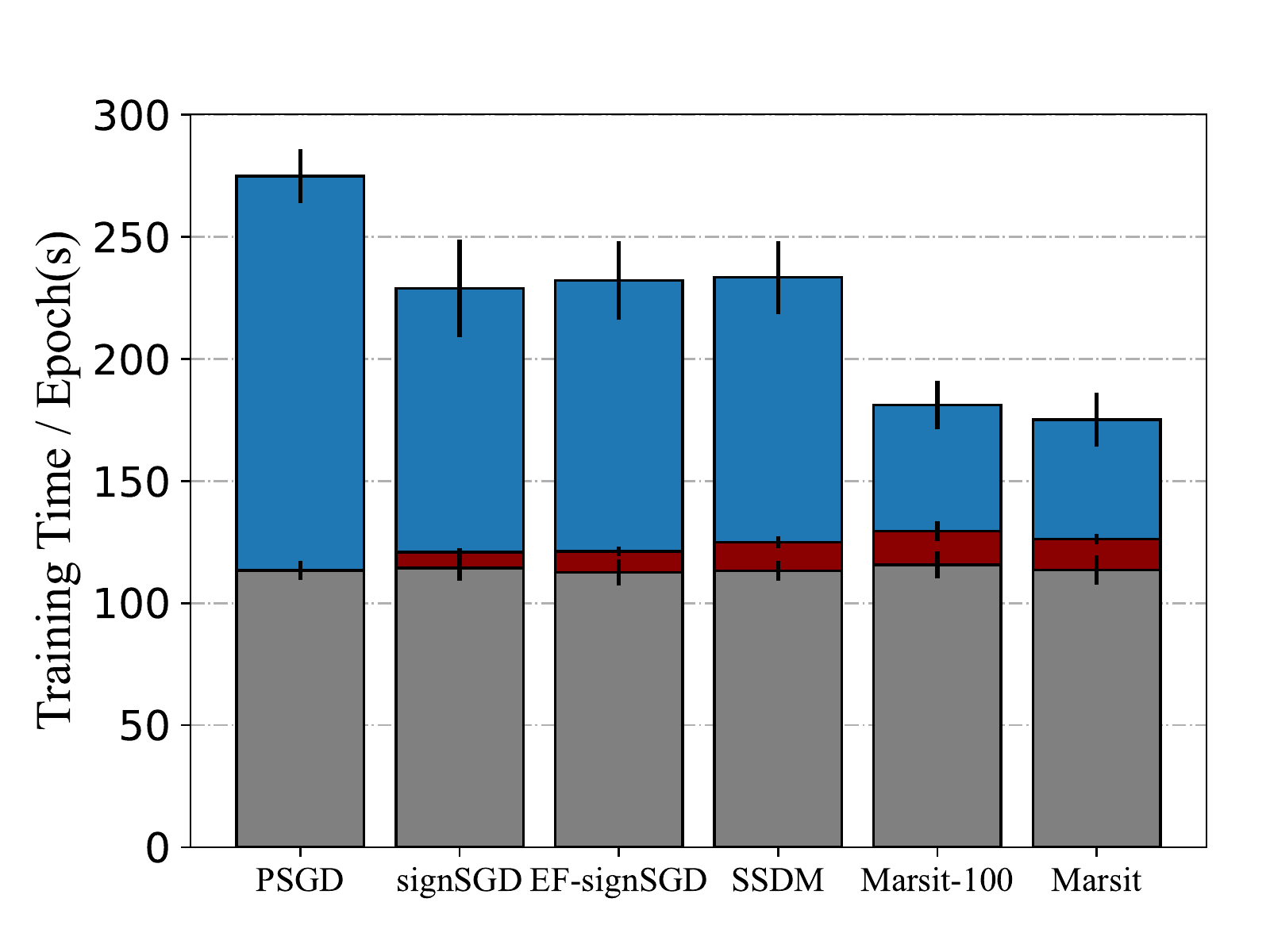}
}%
\centering
\vspace{-10pt}
\caption{Experiments on training AlexNet for CIFAR-10 under TAR and RAR}
\label{fig:exp_speedup}
\end{figure}

\paragraph{Performance under Various MAR settings} Figure \ref{fig:exp_speedup} presents the results of Marsit and its baselines under RAR and TAR. For each method, we measure its average training time in each communication round and split the time into three phases, namely, computation (grey), compression (red) and communication (blue). We notice that Marsit introduces minor compression overheads to prepare for the real-time aggregation. Among these six approaches, it is clear that Marsit and/or Marsit-100 spends the least time in communication compared with other baselines. For TAR paradigm, each baseline takes less time to communicate. For RAR paradigm, the communication time dominates the computation time and Marsit requires less training time between two successive synchronizations.   

\section{Conclusion} \label{sec:conclusion}

This paper proposes a synchronization framework, Marsit, that achieves one-bit transmission under multi-hop all-reduce. In this framework, we design a bit-wise operation to support the receiving and the compression undertake simultaneously. Besides, we introduce a global compensation mechanism to mitigate the compression deviation. Based on the structure, we offer a theoretical guarantee that it achieves the same convergence rate as the non-compression approach using the optimizer of SGD. Empirical studies present that our proposed approach can achieve a similar test accuracy to the non-compression version while using less training time by 35\%. 


\section*{Acknowledgements}

This research was supported by the funding from the Key-Area Research and Development Program of Guangdong Province (No. 2021B0101400003), Hong Kong RGC Research Impact Fund (RIF) with the Project No. R5060-19, General Research Fund (GRF) with the Project No. 152221/19E, 152203/20E, and 152244/21E, the National Natural Science Foundation of China (61872310, 62102131), Shenzhen Science and Technology Innovation Commission (R2020A045), and Natural Science Foundation of Jiangsu Province (BK20210361).

\bibliographystyle{IEEEtran}
\bibliography{sample-base}

\begin{thebibliography}{10}
\providecommand{\url}[1]{#1}
\csname url@samestyle\endcsname
\providecommand{\newblock}{\relax}
\providecommand{\bibinfo}[2]{#2}
\providecommand{\BIBentrySTDinterwordspacing}{\spaceskip=0pt\relax}
\providecommand{\BIBentryALTinterwordstretchfactor}{4}
\providecommand{\BIBentryALTinterwordspacing}{\spaceskip=\fontdimen2\font plus
\BIBentryALTinterwordstretchfactor\fontdimen3\font minus
  \fontdimen4\font\relax}
\providecommand{\BIBforeignlanguage}[2]{{%
\expandafter\ifx\csname l@#1\endcsname\relax
\typeout{** WARNING: IEEEtran.bst: No hyphenation pattern has been}%
\typeout{** loaded for the language `#1'. Using the pattern for}%
\typeout{** the default language instead.}%
\else
\language=\csname l@#1\endcsname
\fi
#2}}
\providecommand{\BIBdecl}{\relax}
\BIBdecl

\bibitem{perez2020object}
F.~P{\'e}rez-Hern{\'a}ndez, S.~Tabik, A.~Lamas, R.~Olmos, H.~Fujita, and
  F.~Herrera, ``Object detection binary classifiers methodology based on deep
  learning to identify small objects handled similarly: Application in video
  surveillance,'' \emph{Knowledge-Based Systems}, 2020.

\bibitem{roy2021application}
K.~Roy, S.~Debdas, S.~Kundu, S.~Chouhan, S.~Mohanty, and B.~Biswas,
  ``Application of natural language processing in healthcare,''
  \emph{Computational Intelligence and Healthcare Informatics}, 2021.

\bibitem{imagenet}
O.~Russakovsky, J.~Deng, H.~Su, J.~Krause, S.~Satheesh, S.~Ma, Z.~Huang,
  A.~Karpathy, A.~Khosla, M.~Bernstein \emph{et~al.}, ``Imagenet large scale
  visual recognition challenge,'' \emph{International journal of computer
  vision}, 2015.

\bibitem{baidu}
Baidu-Research, ``tensorflow-allreduce,'' [Source Code].
  \url{https://github.com/baidu-research/tensorflow-allreduce}, 2017.

\bibitem{horovod}
A.~Sergeev and M.~Del~Balso, ``Horovod: fast and easy distributed deep learning
  in tensorflow,'' \emph{arXiv preprint arXiv:1802.05799}, 2018.

\bibitem{mikami2018massively}
H.~Mikami, H.~Suganuma, P.~U-chupala, Y.~Tanaka, and Y.~Kageyama, ``Massively
  distributed sgd: Imagenet/resnet-50 training in a flash,'' \emph{arXiv
  preprint arXiv:1811.05233}, 2018.

\bibitem{li2014scaling}
M.~Li, D.~G. Andersen, J.~W. Park, A.~J. Smola, A.~Ahmed, V.~Josifovski,
  J.~Long, E.~J. Shekita, and B.-Y. Su, ``Scaling distributed machine learning
  with the parameter server,'' in \emph{OSDI}, 2014.

\bibitem{lu2021optimal}
Y.~Lu and C.~De~Sa, ``Optimal complexity in decentralized training,'' in
  \emph{ICML}, 2021.

\bibitem{lin2021quasi}
T.~Lin, S.~P. Karimireddy, S.~U. Stich, and M.~Jaggi, ``Quasi-global momentum:
  Accelerating decentralized deep learning on heterogeneous data,'' in
  \emph{ICML}, 2021.

\bibitem{chen2021accelerating}
Y.~Chen, K.~Yuan, Y.~Zhang, P.~Pan, Y.~Xu, and W.~Yin, ``Accelerating gossip
  sgd with periodic global averaging,'' in \emph{ICML}, 2021.

\bibitem{resnet}
K.~He, X.~Zhang, S.~Ren, and J.~Sun, ``Deep residual learning for image
  recognition,'' in \emph{CVPR}, 2016.

\bibitem{brown2020language}
T.~B. Brown, B.~Mann, N.~Ryder, M.~Subbiah, J.~Kaplan, P.~Dhariwal,
  A.~Neelakantan, P.~Shyam, G.~Sastry, A.~Askell \emph{et~al.}, ``Language
  models are few-shot learners,'' in \emph{NeurIPS}, 2020.

\bibitem{bernstein2018signsgd_majority}
J.~Bernstein, J.~Zhao, K.~Azizzadenesheli, and A.~Anandkumar, ``signsgd with
  majority vote is communication efficient and fault tolerant,'' in
  \emph{ICLR}, 2018.

\bibitem{safaryan2021stochastic}
M.~Safaryan and P.~Richt{\'a}rik, ``Stochastic sign descent methods: New
  algorithms and better theory,'' in \emph{ICML}, 2021.

\bibitem{liu2018signsgd}
S.~Liu, P.-Y. Chen, X.~Chen, and M.~Hong, ``signsgd via zeroth-order oracle,''
  in \emph{ICLR}, 2018.

\bibitem{tang20211}
H.~Tang, S.~Gan, A.~A. Awan, S.~Rajbhandari, C.~Li, X.~Lian, J.~Liu, C.~Zhang,
  and Y.~He, ``1-bit adam: Communication efficient large-scale training with
  adam’s convergence speed,'' in \emph{ICML}, 2021.

\bibitem{qsgd_tern_wen}
W.~Wen, C.~Xu, F.~Yan, C.~Wu, Y.~Wang, Y.~Chen, and H.~Li, ``Terngrad: Ternary
  gradients to reduce communication in distributed deep learning,'' in
  \emph{NeurIPS}, 2017.

\bibitem{qsgd}
D.~Alistarh, D.~Grubic, J.~Li, R.~Tomioka, and M.~Vojnovic, ``Qsgd:
  Communication-efficient sgd via gradient quantization and encoding,'' in
  \emph{NeurIPS}, 2017.

\bibitem{qsgd_zip_zhang}
H.~Zhang, J.~Li, K.~Kara, D.~Alistarh, J.~Liu, and C.~Zhang, ``Zipml: Training
  linear models with end-to-end low precision, and a little bit of deep
  learning,'' in \emph{ICML}, 2017.

\bibitem{yu2018gradiveq}
M.~Yu, Z.~Lin, K.~Narra, S.~Li, Y.~Li, N.~S. Kim, A.~Schwing, M.~Annavaram, and
  S.~Avestimehr, ``Gradiveq: Vector quantization for bandwidth-efficient
  gradient aggregation in distributed cnn training,'' in \emph{NeurIPS}, 2018.

\bibitem{bernstein2018signsgd_theorem}
J.~Bernstein, Y.-X. Wang, K.~Azizzadenesheli, and A.~Anandkumar, ``signsgd:
  Compressed optimisation for non-convex problems,'' in \emph{ICML}, 2018.

\bibitem{wangni2018gradient}
J.~Wangni, J.~Wang, J.~Liu, and T.~Zhang, ``Gradient sparsification for
  communication-efficient distributed optimization,'' in \emph{NeurIPS}, 2018.

\bibitem{guo2020tail}
J.~Guo, S.~Hu, W.~Wang, C.~Yao, J.~Han, R.~Li, and Y.~Lu, ``Tail: an automated
  and lightweight gradient compression framework for distributed deep
  learning,'' in \emph{DAC}, 2020.

\bibitem{vogels2019powersgd}
T.~Vogels, S.~P. Karinireddy, and M.~Jaggi, ``Powersgd: Practical low-rank
  gradient compression for distributed optimization,'' \emph{NeurIPS}, 2019.

\bibitem{jia2018highly}
X.~Jia, S.~Song, W.~He, Y.~Wang, H.~Rong, F.~Zhou, L.~Xie, Z.~Guo, Y.~Yang,
  L.~Yu \emph{et~al.}, ``Highly scalable deep learning training system with
  mixed-precision: Training imagenet in four minutes,'' \emph{arXiv preprint
  arXiv:1807.11205}, 2018.

\bibitem{cifar10}
A.~Krizhevsky, G.~Hinton \emph{et~al.}, ``Learning multiple layers of features
  from tiny images,'' 2009.

\bibitem{maas-EtAl:2011:ACL-HLT2011}
A.~L. Maas, R.~E. Daly, P.~T. Pham, D.~Huang, A.~Y. Ng, and C.~Potts,
  ``Learning word vectors for sentiment analysis,'' in \emph{Annual Meeting of
  the Association for Computational Linguistics: Human Language Technologies},
  2011.

\bibitem{krizhevsky2012imagenet}
A.~Krizhevsky, I.~Sutskever, and G.~E. Hinton, ``Imagenet classification with
  deep convolutional neural networks,'' \emph{NeurIPS}, 2012.

\bibitem{sanh2019distilbert}
V.~Sanh, L.~Debut, J.~Chaumond, and T.~Wolf, ``Distilbert, a distilled version
  of bert: smaller, faster, cheaper and lighter,'' \emph{arXiv preprint
  arXiv:1910.01108}, 2019.

\bibitem{karimireddy2019error}
S.~P. Karimireddy, Q.~Rebjock, S.~Stich, and M.~Jaggi, ``Error feedback fixes
  signsgd and other gradient compression schemes,'' in \emph{ICML}, 2019.

\bibitem{elias1975universal}
P.~Elias, ``Universal codeword sets and representations of the integers,''
  \emph{IEEE transactions on information theory}, 1975.

\end{thebibliography}

\newpage \appendix \onecolumn
\section{Proof for Cascading Compression} 

\noindent \textbf{SSDM.} An element $v_j$ in vector $\bm{v}$ is compressed for $\{+1, -1\}$ following the probability that: 
\begin{equation*}
    \tilde{\text{sign}}\left(v_j\right) = \begin{cases}
        +1, & pr = \frac{1}{2} + \frac{v_j}{2\|\bm{v}\|}\\
        -1, & pr = \frac{1}{2} - \frac{v_j}{2\|\bm{v}\|}
    \end{cases}
\end{equation*}
where $\tilde{\text{sign}}(\cdot)$ refers to the compression operator. In such an operation, the expected value for $\tilde{\text{sign}}\left(v_j\right)$ is $v_j / \|\bm{v}\|$. Therefore, $\mathbb{E} \tilde{\text{sign}}(\bm{v}) = \bm{v} / \|\bm{v}\|$.  Since the $\ell_2$-norm $\|\bm{v}\|$ is a constant, SSDM can achieve unbiased update with the gradient $\|\bm{v}\| \cdot \tilde{\text{sign}}(\bm{v})$, which we define as $\mathcal{Q}(\bm{v})$. 

Suppose the gradients calculated by all clients are $s^{(1)}$, ..., $s^{(M)} \in \mathbb{R}^D$. Following lists various model updates, including 
\begin{itemize}
    \setlength\itemsep{.5em}
    \item \textbf{Non-compression approach:} $s_1 \overset{\triangle}{=} \frac{1}{M} \sum_{m=1}^M s^{(m)}$ 
    \item \textbf{SSDM under PS:} $s_2 \overset{\triangle}{=} \frac{1}{M} \sum_{m=1}^M \mathcal{Q}\left(s^{(m)}\right)$
    \item \textbf{SSDM using cascading compression:} $s_3 \overset{\triangle}{=} \frac{1}{M} \underbrace{\mathcal{Q}\left(...\mathcal{Q}\left(s^{(1)}\right) + ... + s^{(M)}\right)}_{M \text{ recursive compressions}}$
\end{itemize}
Since the compressor $\mathcal{Q}$ is unbiased, the equality that $\mathbb{E}(s_2) = \mathbb{E}(s_3) = s_1$ holds. It is universally acknowledged that the update under MAR is equivalent to that under PS. In this part, we aim to evaluate the deviation between the compression and the non-compression results, i.e., $\|s_2 - s_1\|_2^2$ for SSDM under PS and $\|s_3 - s_1\|_2^2$ for SSDM using cascading compression. Prior to analyzing these two bounds, we introduce a assumption that widely adopts in \cite{bernstein2018signsgd_theorem, safaryan2021stochastic}: 

\begin{assumption}[Bounded gradient] \label{assumption:unbiased_est}
For any worker $m \in \{1, ..., M\}$ and vector $\bm{x} \in \mathbb{R}^D$, a scalar $G \geq 0$ satisfies 
    \begin{align*}
        \mathbb{E} \left\|s^{(m)}\right\|^2_2 \leq G^2. 
    \end{align*}
\end{assumption}

Next, we first analyze the upper bound for the deviation under PS paradigm: 
\begin{theorem} \label{theorem: ssdm_ps}
Under Assumption \ref{assumption:unbiased_est}, the upper bound for $\|s_2 - s_1\|_2^2$ is $\mathcal{O}(D G^2)$. 
\begin{proof}
Based on the expression of the variance, 
\begin{equation*}
\begin{split}
    \mathbb{E} \left\|\frac{1}{M} \sum_{m=1}^M \mathcal{Q}\left(s^{(m)}\right) - \frac{1}{M} \sum_{m=1}^M s^{(m)}\right\|_2^2 &= \mathbb{E} \left\|\frac{1}{M} \sum_{m=1}^M \mathcal{Q}\left(s^{(m)}\right)\right\|_2^2 - \mathbb{E} \left\|\frac{1}{M} \sum_{m=1}^M s^{(m)}\right\|_2^2 \leq \mathbb{E} \left\|\frac{1}{M} \sum_{m=1}^M \mathcal{Q}\left(s^{(m)}\right)\right\|_2^2\\
    &\leq \frac{1}{M} \sum_{m=1}^M \left\|s^{(m)}\right\|_2^2 \cdot \left\|\tilde{\text{sign}} \left(s^{(m)}\right)\right\|_2^2 \leq D G^2
\end{split}
\end{equation*}
where the second last inequality is based on Cauchy–Schwarz inequality, and the last inequality follows Assumption \ref{assumption:unbiased_est} and the sign matrix containing $D$ $(+1)$s or $(-1)$s, i.e., $\left\|\tilde{\text{sign}} \left(s^{(m)}\right)\right\|_2^2 = D, \forall m \in \{1, ..., M\}$. 
\end{proof}
\end{theorem}

Then, the following theorem analyzes the boundary of cascading compression: 
\begin{theorem} \label{theorem: ssdm_cc}
Under Assumption \ref{assumption:unbiased_est}, the upper bound for the deviation of cascading compression is 
\begin{equation}
   \|s_3 - s_1\|_2^2 \leq \frac{(2D)^M G^2}{M} 
\end{equation}
\begin{proof}
Similar to Theorem \ref{theorem: ssdm_ps}, we have: 
\begin{equation*}
\begin{split}
    \mathbb{E} \left\|\frac{1}{M} \mathcal{Q}\left(...\mathcal{Q}\left(s^{(1)}\right) + ... + s^{(M)}\right) - \frac{1}{M} \sum_{m=1}^M s^{(m)}\right\|_2^2 &\leq \mathbb{E} \left\|\frac{1}{M} \mathcal{Q}\left(...\mathcal{Q}\left(s^{(1)}\right) + ... + s^{(M)}\right)\right\|_2^2\\
    &\leq \frac{D}{M^2} \mathbb{E} \left\|\mathcal{Q}\left(...\mathcal{Q}\left(s^{(1)}\right) + ... + s^{(M-1)}\right)+s^{(M)}\right\|_2^2\\
    &\leq \frac{2D}{M^2} \mathbb{E} \left\|\mathcal{Q}\left(...\mathcal{Q}\left(s^{(1)}\right) + ... + s^{(M-1)}\right)\right\|_2^2 + \frac{2DG^2}{M^2} \leq \frac{G^2}{M^2} \sum_{m=1}^M (2D)^m \leq \frac{G^2}{M} (2D)^M
\end{split}
\end{equation*}
where the third last inequality is based on $\|a + b\|_2^2 \leq 2\|a\|_2^2 + 2\|b\|_2^2$ and follows Assumption \ref{assumption:unbiased_est}, while the last inequality is based on a common sense that $D \geq 1$. Generally speaking, the dimension of a neural network is far larger than the number of workers, i.e., $D >> M$, and therefore, current bound is tighter than the result $G^2 (2D)^{M+1}/M^2$. 
\end{proof}
\end{theorem}

Obviously, when $M=1$ such that the training under PS paradigms is equivalent to that under cascading compression, they have consistent upper bound. Although both theorems show the upper bound, PS paradigm is unlike cascading compression approach that explodes rapidly with respect to the number of workers $M$. 

\section{Proof for Marsit (Theorem \ref{theo:marsit})}

\newcommand{\y}[1]{\tilde{y}_{#1}}
\newcommand{\x}[1]{\tilde{x}_{#1}}
\newcommand{\inner}[2]{\left\langle#1, #2\right\rangle}
\newcommand{\bracket}[1]{\left(#1\right)}
\newcommand{\norm}[1]{\left\|#1\right\|}
\newcommand{\E}{\mathbb{E}}
\renewcommand{\c}[1]{c_{#1}}
\newcommand{\tg}[1]{\tilde{g}_{#1}}
\newcommand{\g}[2]{g_{#1}^{(#2)}}
\newcommand\numberthis{\addtocounter{equation}{1}\tag{\theequation}}

By constructing an auxiliary array $\{\tilde{y}\}$ such that $\y{t} = \x{t} - \c{t}$, where $\c{t} = \sum_{m=1}^M \c{t}^{(m)} / M$, we analyze its recursive function from the following two aspects: 
\begin{itemize}
    \item $\c{t+1} = 0$: 
    \begin{equation}
        \y{t+1} = \x{t+1} = \x{t} - \frac{1}{M} \sum_{m=1}^M \left(\eta_l \g{t}{m} + \c{t}^{(m)}\right) = \y{t} - \frac{\eta_l}{M} \sum_{m=1}^M \g{t}{m}
    \end{equation}
    \item $\c{t+1} \neq 0$: 
    \begin{equation}
        \y{t+1} = \x{t+1} - \c{t+1} = \x{t} - g_t - \frac{1}{M} \sum_{m=1}^M \left(\eta_l \g{t}{m} + \c{t}^{(m)} - g_t\right)  = \y{t} - \frac{\eta_l}{M} \sum_{m=1}^M \g{t}{m}
    \end{equation}
\end{itemize}
Let $\tg{t} = \sum_{m=1}^M \g{t}{m} / M$ and $\g{t}{m}$ here only means $\nabla f_m \bracket{\x{t}; \xi_k^{(m)}}$ in this proof. Obviously, regardless the value of $\c{t+1}$, the recursive function is $\y{t+1} = \y{t} - \eta_l \tg{t}$. According to L-smooth assumption for the non-convex objectives, we have:
\begin{align*}
    \mathbb{E} F(\y{t+1}) - F(\y{t}) &\leq \E\inner{\nabla F \bracket{\y{t}}}{\y{t+1} - \y{t}} + \frac{L}{2} \E\norm{\y{t+1} - \y{t}}_2^2 \\
    &= -\eta_l \E\inner{\nabla F \bracket{\y{t}}}{\nabla F \bracket{\x{t}}} + \frac{L \eta_l^2}{2} \E\norm{\frac{1}{M} \sum_{m=1}^M \g{t}{m}}_2^2 \\
    &\leq -\frac{\eta_l}{2} \bracket{1 - L \eta_l} \norm{\nabla F\bracket{\x{t}}}_2^2 + \frac{\eta_l L^2}{2} \E\norm{\y{t} - \x{t}}_2^2 + \frac{L \eta_l^2 \sigma^2}{2M} \numberthis \label{eq:6}
\end{align*}
where the last inequality is based on the unbiased estimator for the calculated gradient, i.e., $\nabla f_m \bracket{\x{t}; \xi_k^{(m)}}$. Next, we will find the bound for $\E\norm{\y{t} - \x{t}}_2^2$, which is equivalent to $\E\norm{\c{t}}_2^2$. Algorithm \ref{algo:marsit} performs the full precision synchronization every $K$ rounds and therefore, there exists a $t_0 > t - K$ such that $c_{t_0} = 0$. Following analyzes the case that $c_t$ is a non-zero vector: 
\begin{align*}
    \E\norm{\c{t}}_2^2 &= \E\norm{\c{t-1} + \eta_l \tg{t-1} - \eta_s g_{t-1}}_2^2 \\
    &\leq \bracket{1 + \frac{1}{K}} \E \norm{\c{t-1}}_2^2 + (1+K) \E \norm{\eta_l \tg{t-1} - \eta_s g_{t-1}}_2^2 \\
    &\leq \sum_{\tau = t_0}^{t-1} \bracket{1 + \frac{1}{K}}^{t-1-\tau} \cdot (1 + K) \E \norm{\eta_l \tg{\tau} - \eta_s g_{\tau}}_2^2 \\ 
    &\leq 3 (1 + K) \cdot \sum_{\tau = t_0}^{t-1} \E \norm{\eta_l \tg{\tau} - \eta_s g_{\tau}}_2^2 \\
    &\leq 6 \eta_l^2 (1+K) \sum_{\tau = t_0}^{t-1} \E\norm{\tg{\tau}}_2^2 + 6 \eta_s^2 (1+K) \sum_{\tau = t_0}^{t-1} \E\norm{g_{\tau}}_2^2 \\
    &= 6 \eta_l^2 (1+K) \sum_{\tau = t_0}^{t-1} \E\norm{\nabla F(\x{\tau})}_2^2 + 6 \eta_l^2 (1+K)K \cdot \frac{\sigma^2}{M} + 6 \eta_s^2 (1+K) K D \numberthis \label{eq:7}
\end{align*}
where the first inequality is based on $(a+b)^2 \leq (1+\frac{1}{K}) a^2 + (1+K) b^2$, and the last equality is according to $\|g_\tau\|_2^2 = D$ because it is only constituted with $\{+1, -1\}$ for all $D$ dimensions. Suppose the optimal solution for the non-convex objective $F(\cdot)$ is $F_*$. Therefore, plugging the result from Equation \ref{eq:7} into Equation \ref{eq:6}, and summing Equation \ref{eq:6} for all $t$s from 0 to $T$, we have: 
\begin{align*}
    F_* - F(\x{0}) &\leq \sum_{t=0}^{T-1} \bracket{\E F(\y{t+1}) - F(\y{t})}\\
    &\leq -\frac{\eta_l}{2} \bracket{1 - L \eta_l - 3L^2 \eta_l^2 K(K+1)} \sum_{t=0}^{T-1} \norm{\nabla F\bracket{\x{t}}}_2^2 + \frac{\eta_l L^2 T}{2} \bracket{6 \eta_l^2 (1+K)K \cdot \frac{\sigma^2}{M} + 6 \eta_s^2 (1+K) K D} + \frac{L \eta_l^2 \sigma^2 T}{2M}
\end{align*}
By setting $\eta_l = \sqrt{M/T}$ and $\eta_s = 1/\sqrt{TD}$, and assuming that $T$ is sufficiently large, i.e., $T \geq 9 L^2 K^2 (K+1)^2$, we can obtain the desired conclusion. 

\end{document}